\documentclass{IS2023}

\usepackage{multirow}
\usepackage[table,xcdraw]{xcolor}

\IScameraready 

\title{Quantifying and Reducing Bias in E2E Speech Recognition Systems}
\title{Using Data Augmentation and VTLN to Reduce Bias in Dutch End-to-End ASR}
\title{Using Data Augmentations and VTLN to Reduce Bias in Dutch End-to-End Speech Recognition Systems}
\name{Tanvina Patel and Odette Scharenborg}
\address{
  Multimedia Computing Group, Delft University of Technology, Delft, The Netherlands}
\email{t.b.patel@tudelft.nl, O.E.Scharenborg@tudelft.nl}

\begin{document}

\maketitle
 
\begin{abstract}
Speech technology has improved greatly for \textit{norm} speakers,
i.e., adult native speakers of a language without speech impediments or strong accents.
However, \textit{non-norm} or \textit{diverse} speaker groups show a distinct performance gap with norm speakers, which we refer to as bias. In this work, we aim to reduce bias against different age groups and non-native speakers of Dutch. For an end-to-end (E2E) ASR system, we use state-of-the-art speed perturbation and spectral augmentation as data augmentation techniques and explore Vocal Tract Length Normalization (VTLN) to normalise for spectral differences due to differences in anatomy. 
The combination of data augmentation and VTLN reduced the average WER and bias across various diverse speaker groups by 6.9\% and 3.9\%, respectively.
The VTLN model trained on Dutch was also effective in improving performance of Mandarin Chinese child speech, thus, showing generalisability across languages.  
\end{abstract}
\noindent\textbf{Index Terms}: E2E ASR, Bias, Vocal Tract Length Normalization (VTLN), speed perturbations, Spectral augmentations

\section{Introduction}
\label{sec:intro}
Several studies have shown that State-of-the-Art (SotA) Automatic Speech Recognition (ASR) systems struggle with large acoustic variation in speech \cite{Counterfairasr,feng2021quantifying}. These variations can be due to many (demographic) factors, including age \cite{feng2021quantifying}, gender \cite{tatman-2017-genderbias, Garnerin2019GenderRI}, race \cite{RacialDisparityPNAS}, accents \cite{AccentGap}, whispered speech \cite{WhisperSpeechComm}, speech impairment \cite{DeRussis2019OnTIDysarthic}, etc. 
In short, ASR systems perform well for norm speakers, i.e., adult native speakers of a language without speech impediments or strong accents, but show a \textit{bias} against speech from diverse speakers, i.e., those speakers that deviate from the norm. In this work, we analyse and aim to reduce the bias against speakers of different age groups (children, teenagers, adults, older adults) and non-native speakers of Dutch.

An often-mentioned potential source of bias is scarcity of training data from diverse speaker groups. Hence, a potential bias mitigation approach is then generating synthetic training data to reduce the bias against certain speaker groups \cite{prananta22IS,zhang22nIS}. A second potential source of bias are the feature representations \cite{WiebkeFacct22}. Acoustic differences between different age groups are mostly due to differences in vocal tract anatomy \cite{Funggenderchild2021}, while non-native speech is mostly characterised by a noticeable first language (L1) accent in the pronunciation of the second language sounds (L2) \cite{flege1987production}. These acoustic differences between norm and diverse speech may lead to mismatches between the feature representations of norm speech vs. diverse speech, potentially causing performance degradation and bias against diverse speech. 
Here, we aim to improve recognition performance and reduce bias against diverse speech by 1) using SotA data augmentation techniques, specifically speed perturbation \cite{ko2015audioaug} and spectral augmentation \cite{Park2019SpecAug} and, 2) Reducing the feature variability between speaker groups by using Vocal Tract Length Normalization (VTLN) to scale or normalize the acoustic features  \cite{Sarkar2010VocalTL,UmeshVTLN2004}. The VTLN approach has been extensively used to reduce inter-speaker variability for various tasks, e.g., speaker recognition \cite{Sarkar2010VocalTL} and child speech recognition \cite{Gurunath2014ImprovingSpeechRecognitionfor} but mostly in hybrid ASR systems. Since, End-to-End (E2E) ASR systems generally outperform hybrid models for different types of speech, e.g., spontaneous, telephonic, and noisy speech \cite{RNNvsXfmer}, here, we investigate the usability of VTLN within the E2E frame-work. We train the VTLN model using both norm speech and diverse speech. 
VTLN can easily be trained for new languages, as it requires only audio and no extra annotation. However, collecting diverse speech (from several speaker groups) can be difficult, especially in low-resource scenarios. 
Hence, we explore the effectiveness of the VTLN model across languages. To that end, the VTLN model trained on Dutch is applied to Mandarin Chinese speaker groups. 

In this work, the ASR performance is evaluated in terms of Word Error Rate (WER) and bias. Bias is related to WERs, but an improvement in WER may not always imply a reduction in bias (as bias is evaluated with respect to a certain speaker group).
An important open question is how to actually measure bias. Recently, studies have proposed measures to quantify bias against various speaker groups. In the ASR and speaker recognition literature, bias measures are generally defined as differences or ratios between the base metrics (e.g., WER, EER) of a speaker group and a reference group. 
For e.g., in \cite{feng2021quantifying,zhang22nIS}, bias against a specific diverse speaker group is computed by taking the absolute WER difference with the best performing diverse speaker group. The authors in \cite{CasualConvIcassp22Chunxi, dheram22IS} propose a similar measure but use the relative WER gap as bias measure. Generally, the reference group is the minimum WER group in the category, however, there are some drawbacks to these measures (see Section 3.2) and hence we propose a new bias measure. 
Summarizing, in this work, we investigate the effectiveness of data augmentation and feature normalization (VTLN) as bias mitigation approaches in a Dutch E2E ASR system, focusing on both read and conversational speech, and propose a new bias measure.

    
\section{Methodology}
\label{sec:methology}
Here, we describe the process of data augmentation and feature normalization by VTLN used for E2E training.

\subsection{Data Augmentation}
We consider two types of data augmentations: one applied to the raw audio wave file and one to the feature vector, i.e., speed perturbations \cite{ko2015audioaug} to increase the training data and spectral augmentation to improve system robustness \cite{Park2019SpecAug}, respectively.

\smallskip
\noindent \textit{Speed Perturbation (SP):} Speed perturbation is performed by resampling the original raw speech signal which results in a warped time signal. Given an audio speech signal $s(t)$, time warping by a factor $\beta$ gives the signal $s(\beta t)$. The Fourier transform of $s(\beta t)$ is $S(\omega/\beta)/\beta$. This implies that, in addition to the change in the duration of the signal which affects the number of frames in the utterance, the warping factor produces shifts in the frequency components (shift of the speech spectrum). Adding speed perturbed data to the training data has shown to improve ASR recognition performance \cite{ko2015audioaug}.

\smallskip
\noindent \textit{Spectral Augmentation (SpecAug)}: 
Spectral Augmentation is applied on the log mel spectrogram of the input speech rather than the raw waveform itself. It consists of three augmentation policies: 1) time masking and 2) frequency masking (that masks a block of consecutive time steps or mel frequency channels) and 3) time-warping that randomly warps the spectrogram along the time axis. SpecAug does not increase or reduce the duration of the speech signal but squeezes and stretches the spectrogram locally. Using SpecAug is computationally efficient and has also shown to improve ASR recognition performance \cite{Park2019SpecAug}.

\subsection{Vocal Tract Length Normalization (VTLN)}
The vocal tract length varies from person to person and across age groups leading to variations in the speech spectrum due to the formants shifting in frequency in an approximately linear fashion. The process of compensating spectral variation due to vocal tract length variation is known as Vocal Tract Length Normalization (VTLN). The process of VTLN includes: 

\begin{enumerate}[leftmargin=*]
    \item Train a VTLN model on a given speech database. 
    \item Estimate the warping factor $\alpha$ for a given test utterance and normalize the features of the test utterance with the factor.
\end{enumerate}

\noindent The process of VTLN warps the features to that of an ideal or reference speaker ($\alpha_r=1$). For adult, male speakers, the energy in the speech spectrum is towards the lower frequencies, while it is higher for females, hence, their estimated warping factors are around $\alpha_m\geq\alpha_r$ and $\alpha_f\leq\alpha_r$, respectively. For children, since their spectrum energies are typically even higher than female speakers, it is expected that  $\alpha_c<\alpha_r$ to compress the frequency axis closer to the reference. The VTLN model training is done as in \cite{kaldivtln}, which uses a linear feature transform corresponding to each warp factor  \cite{UmeshVTLN2004} with a grid search that finds out the best $\alpha$ in the range $[0.80,1.20]$. 



\section{Experimental setup}
\subsection{Databases}
We consider two Dutch databases: the Corpus Gesproken Nederlands (CGN) \cite{oostdijk2000spoken} for training the ASR system and the Jasmin-CGN corpus \cite{cucchiarini2006jasmin} for testing the different speaker groups. Additionally, we use the Mandarin Chinese Spoken Language Technology (SLT) 2021 database \cite{Yu2021SLT} for investigating the language-independence of the VTLN model trained on Dutch language.

\subsubsection{The Dutch Corpora}

\textit{Corpus Gesproken Nederlands (CGN)} \cite{oostdijk2000spoken}: The corpus consists of native speech data spoken by norm speakers within the 18-65 years age range from the Netherlands and Flanders. We use the Netherlands data consisting of monologue and multilogue speech. The data includes lecture recordings, broadcast data, spontaneous conversations, telephonic speech, etc. The unprocessed training data consists of around 480 hours of speech and the CGN test data consists of read broadcast news (Rd) and conversational telephone speech (CTS). Table \ref{Table:CGN-Jas-Mix} shows the train, development, and test partitions, as in \cite{leeuwen09bIS}

\noindent\textit{Jasmin corpus} \cite{cucchiarini2006jasmin}: This corpus is an extension of the CGN corpus\footnote{CGN and Jasmin are recorded under a variety of conditions (potentially non-overlapping) leading to potentially mismatched scenarios.} 
consisting of read speech and Human Machine Interaction (HMI) speech spoken by various diverse speaker groups, i.e., native and non-native speaking children, teenagers and older adults, see Table \ref{Table:CGN-Jas-Mix} for an overview. 
\begin{table}[h]
\caption{Details of the Dutch CGN and Jasmin CGN database}
\label{Table:CGN-Jas-Mix}
\resizebox{\linewidth}{!}{
\small
\centering
    \begin{tabular}{ccccccc}
        \hline
        \textbf{Dataset} & \textbf{Style} & \textbf{Spks}& \multicolumn{4}{c}{\textbf{Hours}} \\
         &  &  & Train & Dev & Test-Rd & Test-CTS \\ 
        \hline
        CGN         & Read $|$ CTS & 2897 &  433 & 43 & 0.45 & 1.80 \\
      \hline 
    \end{tabular}}
\end{table}
\vspace{-0.7cm}
\begin{table}[h]
\label{Table:CGN|Jasmin}
\resizebox{\linewidth}{!}{
\small
\centering
    \begin{tabular}{lccccc}
        \hline
        \textbf{Dataset: Jasmin} & \textbf{Style} & \textbf{Age} & \textbf{Spks}& \multicolumn{2}{c}{\textbf{Hours}} \\
         &  &  & & Read & HMI \\ 
        \hline 
        Native Children: DC  & Read $|$ HMI & 6-13   & 71 & 6.55 & 1.55\\
        Native Teenagers: DT  & Read $|$ HMI & 12-18  & 63 & 4.90 & 0.94\\
        Non-native Teenagers: NnT & Read $|$ HMI & 11-18  & 53 &  6.03 & 1.16\\
        Non-native Adults: NnA & Read $|$ HMI & 19-55  & 45 &  6.01 & 3.07 \\
        Native OlderAdults: DOA & Read $|$ HMI & 65+  &  68 & 6.38  & 3.89\\       
     \hline 
    \end{tabular}}
    \vspace{-0.3cm}
\end{table}

\subsubsection{The Mandarin Database}
This dataset is a part of the Children Speech Recognition Challenge at the IEEE SLT 2021 workshop \cite{Yu2021SLT}. It has different aged speaker groups, and thus, will allow us to study the language-independence of the VTLN model trained on Dutch. The Sets A, C1, and C2 consist of adult read speech, child read speech and child conversational speech, respectively. Table \ref{SLT:Mandarain} shows training, development and test sets as in \cite{ng2020cuhk}. 
\begin{table}[h]
\caption{Details of the Mandarin SLT database}
\label{SLT:Mandarain}
\resizebox{\linewidth}{!}{
\small
\centering
    \begin{tabular}{cccc|ccc|c}
        \hline
        \textbf{Set} & \textbf{Style} & \textbf{Age} & \textbf{Spks} & \multicolumn{3}{c|}{\textbf{Hours}} & \textbf{Total} \\ 
         &  &  & & Training & Dev & Test & Hours \\ \hline        
        A & Read & 18-60 & 1999 & 276.7 & 31.52 & 33.41  & 341 \\
        C1 & Read & 7-11 & 927 & 23.38 & 2.48 & 2.79 & 29 \\
        C2 & Conv. &  4-11 & 166 & 23.49 & 2.85 & 3.14 & 30 \\
     \hline
     \end{tabular}}
     \vspace{-0.2cm}
\end{table}

\begin{table*}[h]
\caption{Results in \%WER for the Dutch ASR system when trained on CGN and tested on CGN and Jasmin. SP = for Speed Perturbation.}
\label{Table:DutchResults}
\resizebox{\linewidth}{!}{
\begin{tabular}{@{}clc|cc|ccccc|ccccc|c@{}}
\hline 
\textbf{}      & \textbf{}      & \textbf{}     & \multicolumn{2}{c|}{\textbf{CGN}} & \multicolumn{5}{c}{\textbf{Jasmin: Read}}   & \multicolumn{5}{c}{\textbf{Jasmin: HMI}} & \textbf{Jasmin} \\ \hline

\textbf{Training} & \textbf{Augmentation} & \textbf{Normalization} & \textbf{Rd}    & \textbf{CTS} & \textbf{DC} & \textbf{DT} & \textbf{NnT} & \textbf{NnA} & \textbf{DOA} & \textbf{DC} & \textbf{DT} & \textbf{NnT} & \textbf{NnA} & \textbf{DOA} &\textbf{Avg.} \\
\hline
       &(a) None   & None  & 9.6 & 23.9      & 42.9  & 22.1        & 54.0 & 59.0 & 28.1 &  50.2        & 40.1        & 59.9 & 60.6 & 41.8 & 45.87  \\
       
       &(b) SP   & None   & 7.0 & 22.0 & 36.7        & 20.5        & 55.6 & 61.2 & 27.2 & 43.8  & 35.4  & 60.3 & 60.8   & 41.2 & 44.27 \\
       
       &(c) SP + SpecAug   & None  & \textbf{7.0} & 20.2			& 36.1        & 18.8        & 51.1 & 58.8 & 26.0 &  40.1        & 27.8        & 52.6 & 55.9 & 38.0 & 40.52\\
       
       \cmidrule(l){2-16} 
       &(d) None       & VTLN$_\text{CGN}$  & 9.3   & 23.6    & 38.8        & 21.2        & 53.4 & 58.3 & 27.2  & 45.9        & 34.9        & 59 & 61.1 & 41.5 & 44.13\\
       &(e) None        & VTLN$_\text{Jasmin}$  & 9.3 & 24.2       & 37.5        & 23.0        & 55.8 & 60.2 & 29.6 & 44.4        & 38.3        & 60.5 & 61.5 & 42.4 & 45.32\\
       \cmidrule(l){2-16} 
       &(f) SP + SpecAug    & VTLN$_\text{CGN}$  & 7.3 & \textbf{20.2} & 34.0 & 17.9  & 50.5 & 56.6 & 24.1 & \textbf{37.5} & \textbf{27.4} & \textbf{52.2} & 55.1 & \textbf{35.4} & 39.07\\
\multirow{-8}{*}{\rotatebox{90}{CGN: Adult Speech}} &(g) SP + SpecAug        & VTLN$_\text{Jasmin}$  & 7.2   & 20.3     & \textbf{32.6}        & \textbf{17.8}        & \textbf{49.8} & \textbf{55.4} & \textbf{23.7} & 37.7  & 29.0  & 52.4 & \textbf{54.7}   & 36.4 & \textbf{38.95} \\ 
\bottomrule
\vspace{-0.9cm}
\end{tabular}}

\end{table*}
\subsection{ASR System Architecture}
For our ASR experiments, we use the conformer architecture \cite{ConformerAnmolG} trained using the ESPNet toolkit \cite{DBLPESPnet}. The other features and training parameters are as follows:

\smallskip 
\noindent \textit{Features}: The front-end features are $80$ dimensional log-mel filterbank features with $3$-dimensional pitch features used for network training. The audio files are sampled at 16kHz.


\smallskip 
\noindent\textit{Dictionary}: For the Dutch ASR system, a unigram model with 5000 byte pair tokens is used. For the Mandarin ASR, a character level model is build with 5767 characters.

\smallskip 
\noindent\textit{Augmentation parameters}: The training data is perturbed by modifying the speed to 90\% and 110\% of the original rate creating a 3-fold training set. Post speed perturbation, SpecAug is used with default settings within, maximum width of each time and frequency mask, $T=40$, $F=30$, respectively.

\smallskip 
\noindent\textit{Normalization}: The MFCC features are used to train a VTLN model using the kaldi recipe \cite{kaldivtlncode}. For each wave file, the VTLN model estimates a single warping factor typically in the range 0.8 to 1.2. The warping factors are used to scale the frequency axis during front-end feature extraction. 
The VTLN model is trained on two different datasets, VTLN$_\text{CGN}$: trained on norm speech (CGN) and VTLN$_\text{Jasmin}$: trained on diverse speech (Jasmin). This allows us to investigate the effect of training on norm vs. diverse speech on the estimated warping factors and ASR performance (Section 4.2).

\smallskip 
\noindent\textit{Evaluation (Error Rate)}: We use the Word Error Rate (WER) and Character Error Rate (CER) to evaluate the Dutch and Mandarin ASR systems performance, respectively. 

\smallskip
\noindent\textit{Evaluation (Bias)}: 
Generally, bias of the diverse speaker group is estimated w.r.t a reference speaker group. The reference group is for instance the minimum WER group in the category \cite{feng2021quantifying,dheram22IS}, however, this means that the bias of the reference group itself cannot be estimated. Also, a minimum WER group may not always exist. Hence, we consider the norm group as the reference speaker group.
If $WER_{norm}$ is the WER of the norm group
of speakers and $WER_{spk_g}$ is the WER of the diverse speaker group ${spk_g}$ (assuming $WER_{spk_g} > WER_{norm}$) then the \textit{Individual Bias} for speaker group ${spk_g}$ is, 
\begin{align}
\mathnormal{Individual Bias} = WER_{spk_g}-WER_{norm}
  \end{align}
Thus, for a total of $G$ speaker groups, the \textit{Overall Bias} of the system can be defined as, 
\begin{equation}
    \mathnormal{Overall Bias} = 1/G \sum_{g} WER_{spk_g}-WER_{norm} .
\end{equation}
Here, $G=10$, when estimating the overall ASR system bias, i.e., five diverse speaker groups for read and HMI each.
\section{Results and Discussions}
We investigate the effect of data augmentation techniques and VTLN separately and combined. 
Table \ref{Table:DutchResults} presents the WERs for different speaker groups and different speaking styles.

\label{sec:Results&Discussions}
\subsection{Baseline ASR}
The baseline ASR system (no augmentation or normalization; row a) achieves 9.6\% and 23.9\% WER on read and continuous speech for norm speakers of CGN (matched condition), respectively. The baseline performed (much) worse on the Jasmin speaker groups, with the worst performances for non-native adults (NnA) and teens (NnT), and native children (DC). Even the better recognised diverse speaker groups have WERs that are more than twice that of the norm speaker group.

\subsection{Experiments related to data augmentation and VTLN}

\noindent\textit{Effect of Data Augmentation}: Adding data using speed perturbations improves performance for the norm and diverse native speaker groups (row b). The improvement is largest in DC, thus, time compression and frequency scaling using SP seems to benefit child speech recognition the most. A slight performance degradation is observed for the non-native speakers, which is expected as with SP, the amount of native (norm) data is increased thus (further) skewing the norm vs. diverse speech distribution in the training data. SpecAug improves recognition performance for the non-native speakers, mostly for HMI speech (row c). Averaged over all speaker groups, adding both SP and SpecAug decreases the WER by $\sim$3\% and $\sim$7\% for read and HMI speech, respectively, compared to baseline.

\begin{figure}[h]
  \centering
  \includegraphics[trim={0.1cm 0.2cm 0cm 0cm}, clip, width=\linewidth]{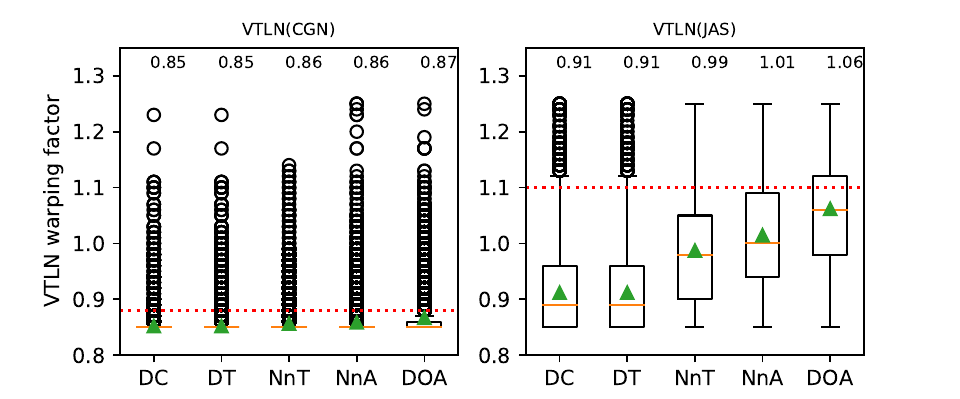}
  \caption{Warping factors estimated across speaker groups from different VTLN models (averaged over speaking styles). Red dotted line: warping factor for norm (adult CGN) speakers.}
  \label{fig:VTLN-warps-boxplot}
\end{figure}

\smallskip
\noindent\textit{Effect of VTLN}: 
\noindent We investigate the warping factors estimated for each of the test speaker groups by the two different VTLN models by visualising them in the box plots in Fig. \ref{fig:VTLN-warps-boxplot}. With the VTLN$_\text{CGN}$, almost all speaker groups have $\alpha<0.9$. This may be due to the fact that the model is trained with only adult speech from CGN. However, when the VTLN model is trained on diverse speech, VTLN$_\text{Jasmin}$, which includes almost equal amounts of data from different age groups, the warping factors are estimated well (child speech $\alpha<1$ and adult speech $\alpha\approx1$) \cite{ghaisinha2016}.Why these better warping factors did not lead to better performance than VTLN$_\text{CGN}$ is a topic for further investigation.


\vspace{-0.2cm} 

\noindent The effect of VTLN on ASR system performnce is shown in Table \ref{Table:DutchResults} (row d,e). With VTLN$_\text{CGN}$, the WER is lower than the baseline for almost all speaker groups. For VTLN$_\text{Jasmin}$, the results are mixed and only significant improvement is seen for child speech (row e). Using VTLN performs slightly better than baseline and similar (or a bit worse) than when using data augmentation even though data augmentation leads to thrice the amount of training data compared to when using VTLN.

\smallskip
\noindent\textit{Effect of augmentation and VTLN}:
To investigate the effect of data augmentation and VTLN together, we apply both SP and SpecAug and also normalize the features while training the models. With VTLN model trained on CGN and Jasmin (rows f and g), respectively, the performance improved across all speaker groups compared to when only using augmentations (row-c), indicating that the bias reduction methods are complementary in their effect on the WER. 
In addition, the better warping factors as estimated with VTLN$_\text{Jasmin}$, (see Fig. 1) indeed lead to the lowest average WER of all systems. The performance improvement for the diverse speaker groups is observed without much affecting the performance of norm speakers.

\subsection{Bias in the Dutch ASR System}
Table \ref{Table:DutchBiasResults} shows the bias as calculated using the WERs in Table \ref{Table:DutchResults}. The overall bias is larger for read speech than for HMI speech for all models. 
This is most likely due to the very low WER for norm CGN read speech (Rd) compared to norm CGN conversational speech (CTS), thus resulting in a larger WER gap and a larger bias against diverse speakers for read speech than HMI speech.
The average overall bias, reduced by 2.2\%  with SP+SPecAug compared to baseline. And on further applying VTLN, the bias reduced by an additional 1.72\%.

\begin{table}[h]
\caption{Overall Bias for the Dutch ASR system (darker cells represents relatively more bias than the norm speech)}
\label{Table:DutchBiasResults}
\centering
\scalebox{0.85}{
\begin{tabular}{ccccc}
\toprule
\multicolumn{1}{l}{\textbf{}} & \multicolumn{1}{l}{\textbf{}} & \multicolumn{3}{c}{\textbf{Overall Bias}} \\
\cmidrule{3-5}
\textbf{Augmentation}  & \textbf{Normalization} & \textbf{Read} & \textbf{HMI}  & \textbf{Average}  \\
\midrule

None & None & \cellcolor[HTML]{EA8E87}31.62 & \cellcolor[HTML]{F4C5C1}26.62 & \cellcolor[HTML]{EFAAA4}29.12 \\
SP & None & \cellcolor[HTML]{E67C73}33.24 & \cellcolor[HTML]{F5C9C5}26.3 & \cellcolor[HTML]{EEA39C}29.77 \\
SP + SpecAug & None & \cellcolor[HTML]{EB938C}31.16 & 
\cellcolor[HTML]{FDF1F0}22.68 & \cellcolor[HTML]{F4C2BE}26.92 \\
\midrule
None & VTLN$_\text{CGN}$ & \cellcolor[HTML]{EC9B94}30.48 & \cellcolor[HTML]{F8D8D6}24.88 & \cellcolor[HTML]{F2BAB5}27.68 \\
None & VTLN$_\text{Jasmin}$ & \cellcolor[HTML]{E98B83}31.92 & \cellcolor[HTML]{F7D5D2}25.22 & \cellcolor[HTML]{F0B0AA}28.57 \\
\midrule
SP + SpecAug & VTLN$_\text{CGN}$ & \cellcolor[HTML]{EFA8A2}29.32 & \cellcolor[HTML]{FFFFFF}21.32 & \cellcolor[HTML]{F7D4D1}25.32 \\
SP + SpecAug & VTLN$_\text{Jasmin}$ & \cellcolor[HTML]{F0AFA9}28.66 & \cellcolor[HTML]{FFFBFB}21.74 & \cellcolor[HTML]{F7D5D2}25.20 \\ \bottomrule
\end{tabular}}
\end{table}

\noindent Figure 2 shows the average bias for the individual diverse speaker groups for the baseline system (blue), when applying data augmentations (red), VTLN trained on Jasmin (yellow) and when applying both (green). The bias was largest for NnA, NnT, DC, DOA, DT in order of decreasing bias. Importantly, the best performing system, i.e., with data augmentation and VTLN trained on Jasmin, also resulted in the lowest bias for all diverse speaker groups. 
The smallest bias for native teenagers can potentially be due to their vocal tract characteristics and speaking styles being similar to those of norm speakers, while the vocal tract characteristics of children and the speaking styles of non-native speakers and older adults differ (vary) more from norm speech, negatively impacting recognition performance. 
 
\begin{figure}[h]
  \centering
  \includegraphics[trim={0cm 0cm 0cm 0cm}, clip, width=\linewidth]{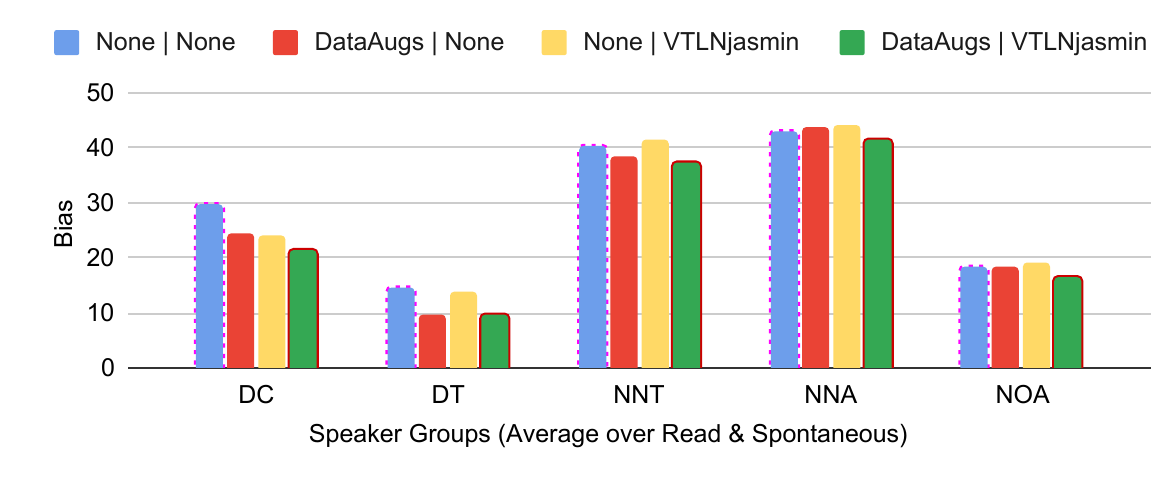}
  \caption{The bias estimated for the diverse speaker groups for different models (Model: Augmentation $|$ VTLN)}
  \label{fig:VTLN-warps}
\end{figure}
\vspace{-0.2cm}

\subsection{Language Independence of the VTLN Model}
To investigate whether the VTLN model can be used across languages we used the two VTLN models trained on Dutch to estimate the warping factors for the Mandarin Chinese speaker group during testing.  The baseline Mandarin ASR system is trained using the Mandarin adult read speech data from SetA (norm speech), with speed perturbations and SpecAugment (similar to the Dutch model). Next, using the VTLN models (VTLN$_\text{CGN}$ and VTLN$_\text{Jasmin}$), we estimate the warping factors for the test sets SetA (norm), SetC1 (child read speech), SetC2 (child spontaneous speech) of the Mandarin dataset. 

\noindent Table \ref{Table:MandariwhnResultsonlyTestVTLN} (row-a) shows the CERs for the baseline system without normalization, and when VTLN$_\text{CGN}$ (row b) and VTLN$_\text{Jasmin}$ (row c) VTLN models are applied to the test sets. 
For the baseline, the CER for child read speech (SetC1) is highly similar to that of the adult speakers (SetA). The performance for the conversational speech of 4-11 year old children (SetC2) is almost 4 times higher than norm speech, likely due to the younger age of some of the speakers and of course due to the conversational nature of the speech. 
Considering that SetA consists only of adult (norm) speech, we did not expect to find an improvement for SetA, which was indeed the case. Despite expecting improvements for the two child speech sets, none was observed for the SetC1. For the conversational child speech (SetC2), a small reduction in CER was observed for both the VTLN$_\text{CGN}$ and the   VTLN$_\text{Jasmin}$ models. In short, we observe that feature normalization by VTLN can help to reduce the pronunciation variations due to vocal tract differences across languages. 

\begin{table}[t]
\caption{Results in \%CER for the Mandarin ASR system when tested with and without VTLN models trained on Dutch}
\label{Table:MandariwhnResultsonlyTestVTLN}
\centering
\scalebox{1.0}{
\small
\centering
\begin{tabular}{cccccc}
\toprule
\textbf{Training} & \textbf{Normalization} &\textbf{SetA} & \textbf{SetC1} & \textbf{SetC2} \\ 
\midrule
(a) SetA & None & 9.9  & 10.0 &  38.8\\
(b) SetA & VTLN$_\text{CGN}$  & 10.2  & 9.9 &  37.1\\
(c) SetA & VTLN$_\text{Jasmin}$  & 9.9  & 9.9 &  37.3 \\

\bottomrule
\end{tabular}}
\vspace{-0.3cm}
\end{table}
\section{Summary and Conclusions}
In this work, we investigated the effectiveness of using data augmentation and feature normalization by VTLN with E2E models. We observe that with augmentation and VTLN, there is a reduction in WER and in bias against age and non-native accented speech. 
Generally, VTLN has been applied for child speech recognition and in an hybrid ASR framework while in this work,
we investigate the usefulness of VTLN for improving recognition performance and reducing bias against other diverse speaker groups as well in an E2E-ASR framework. 

We observed improved recognition performance when using only SP for the native speaker groups. Adding SpecAug improved the recognition performance of the non-native speakers particularly. Thus, data augmentations helped to use norm speaker data to improve performance of diverse speakers.  VTLN gave comparable recognition results across the board but with far less training data. The combination of speed perturbation, SpecAug, and VTLN gave the best recognition performances and reduced bias the most. 
Bias was and remained highest against non-native speakers, which implies that the acoustic properties of native and non-native accented speakers are rather different and cannot be straightforwardly compensated with data augmentation or feature normalization. 

Ideally the warping factors are speaker specific and should be language independent. Our final experiment showed that a VTLN model trained on one language is able to some extent extract warp factors for another language and hence, VTLN can be used as a pre-processing module to the ASR for another language.
With just normalizing the test features, improvement is observed. Possibly, the VTLN model can be further improved when trained with diverse speech from several languages as well.
In the future, we the efficacy of VTLN and other combinations of data augmentation techniques to further reduce the bias against non-native speakers and improve recognition performance and lower bias across more diverse groups, in our aim to build inclusive automatic speech recognition.


 

\newpage
\bibliographystyle{IEEEtran}
\bibliography{CGN_Jasmin_BiasReduction_ArXiv}

\end{document}